\documentclass[conference]{IEEEtran}
\IEEEoverridecommandlockouts

\usepackage{cite}
\usepackage{amsmath,amssymb,amsfonts}
\usepackage{algorithmic}
\usepackage{graphicx}
\usepackage{textcomp}
\usepackage{xcolor}
\usepackage{ulem}
\usepackage{tabularx, multirow, booktabs}
\def\BibTeX{{\rm B\kern-.05em{\sc i\kern-.025em b}\kern-.08em
    T\kern-.1667em\lower.7ex\hbox{E}\kern-.125emX}}

\usepackage[hidelinks]{hyperref}
\usepackage{multirow}
\usepackage{graphicx}
\usepackage{bbold}
\useunder{\uline}{\ul}{}
\usepackage{booktabs}
\usepackage{multirow}

\usepackage{amsmath}
\usepackage{subfigure}
\usepackage{multirow, tabularx, booktabs}
\usepackage{enumerate} 
\usepackage{siunitx}

\usepackage{fancyhdr}
\fancypagestyle{firstpage}{
  \fancyhf{} 
  
  \fancyfoot[L]{979-8-3503-2445-7/23/\$31.00 ©2023 IEEE} 
}
\title{Stock Movement and Volatility Prediction 
 from Tweets, Macroeconomic Factors
and Historical Prices}

\author{\textbf{Shengkun Wang}$^1$, \textbf{YangXiao Bai}$^2$, \textbf{Taoran Ji}$^3$, \textbf{Kaiqun Fu}$^2$, \textbf{Linhan Wang}$^1$, \textbf{Chang-Tien Lu}$^1$\\
$^1${Department of Computer Science, Virginia Tech} \\
$^2${Department of Computer Science, South Dakota State University} \\
$^3${Department of Computer Science, Texas A\&M University - Corpus Christi} \\

\texttt{shengkun@vt.edu}, \texttt{bai.yangxiao@sdstate.edu}, \texttt{taoran.ji@tamucc.edu},\\ 
 \texttt{kaiqun.fu@sdstate.edu},  \texttt{linhan@vt.edu}, \texttt{ctlu@vt.edu} }

\begin{document}

\maketitle
\begin{abstract}
Predicting stock market is vital for investors and policymakers, acting as a barometer of the economic health. We leverage social media data, a potent source of public sentiment, in tandem with macroeconomic indicators as government-compiled statistics, to refine stock market predictions. However, prior research using tweet data for stock market prediction faces three challenges. First, the quality of tweets varies widely. While many are filled with noise and irrelevant details, only a few genuinely mirror the actual market scenario. Second, solely focusing on the historical data of a particular stock without considering its sector can lead to oversight. Stocks within the same industry often exhibit correlated price behaviors. Lastly, simply forecasting the direction of price movement without assessing its magnitude is of limited value, as the extent of the rise or fall truly determines profitability. In this paper, diverging from the conventional methods, we pioneer an  ECON (A Fram\underline{e}work Leveraging Tweets, Ma\underline{c}roeconomic Indicat\underline{o}rs, and Historical Prices to Predict Stock Moveme\underline{n}t and Volatility). The framework has following advantages: First, ECON has an adept tweets filter that efficiently extracts and decodes the vast array of tweet data. Second, ECON discerns multi-level relationships among stocks, sectors, and macroeconomic factors through a self-aware mechanism in semantic space. Third, ECON offers enhanced accuracy in predicting substantial stock price fluctuations by capitalizing on stock price movement. We showcase the state-of-the-art performance of our proposed model using a dataset, specifically curated by us, for predicting stock market movements and volatility.
\end{abstract}

\thispagestyle{firstpage}

\begin{IEEEkeywords}
stock market prediction, tweets extraction, time series forecasting, sentiment analysis, macroeconomic factors
\end{IEEEkeywords}

\section{Introduction}

The stock market, a vital mechanism for facilitating stock trading and crucial capital raising for companies, exerts a significant influence on other business sectors~\cite{billah2016stock}. 
Over the past few years, the U.S. stock market capitalization-to-GDP ratio\footnoterule\footnote{The ratio is a measure of the total value of all publicly traded stocks in a market divided by that economy's gross domestic product (GDP). } nearly exceeded 200\%, and despite a slight dip in 2023, it has stabilized at around 150\%.

This prominence underscores the stock market's role as a key indicator of the U.S. economy. Blue-chip stocks\footnoterule\footnote{A blue chip stock is stock issued by a large, well-established, financially-sound company with an excellent reputation.}, serving as a microcosm of the stock market, become the focus of our research. Our study on the financial market employs a selection of 42 blue-chip stocks from 10 Global Industry Classification Standard (GICS) \footnoterule\footnote{GICS is a method for assigning companies to a specific economic sector and industry group that best defines its business operations.} Sectors, each deemed as investment-worthy\footnoterule\footnote{Investment-worthy refers to high-quality companies rated Baa or higher according to evaluations by Moody's and Standard \& Poor's (S\&P).} by both Moody's and S\&P. Acknowledging the inherent unpredictability of precise stock prices~\cite{nguyen2015sentiment}, our research leverages these blue-chip stocks to forecast future trends in stock price movement and volatility~\cite{feng2019temporal,xu-cohen-2018-stock}.

Research within the realm of the stock market generally falls into two branches: technical analysis and fundamental analysis. Technical analysis employs historical stock prices as features to forecast future movements\cite{lu2021cnn}. However, it is heavily reliant on past data, and often fails to account for abrupt market shifts triggered by unforeseen events. This methodology also assumes uniform rational market behavior, potentially leading to an echo chamber effect where trading signals self-reinforce, detaching from the underlying economic reality. 
Fundamental analysis, on the other hand, incorporates not only price features but also external information, such as social media data~\cite{cookson2020don} and search engine data~\cite{wang2021stock}. Mao et al.\cite{10.1145/2392622.2392634} shows the accuracy of predicting the S\&P 500 closing price increases when Twitter\footnote{Although Twitter has recently been re-branded to ``X'', this article continues to use the original name, ``Twitter''.} data is incorporated into their model. Commonly, these data sources often serve as valuable indicators, mirroring not just the financial market but also other key economic indicators. However, current fundamental analysis research predominantly focuses on the historical information of individual stock\cite{soun2022accurate}, overlooking the interplay between the macro-economy and the stock market. Moreover, the  existing models are primarily focused on predicting whether trends will change\cite{zhang2022transformer}, but overlook the significance of the magnitude of such changes. In reality, the extent of these changes constitutes an important part of stock behavior.

In this paper, to address the aforementioned limitations, we propose ECON 
(A Fram\underline{e}work Leveraging Tweets, Ma\underline{c}roeconomic Indicat\underline{o}rs, and Historical Prices to Predict Stock Moveme\underline{n}t and Volatility).

Our contributions can be summarized as follows:

\textbf{\textbullet\ 
Designing an efficient framework for tweet data extraction in financial domain.} 
The framework employs a sophisticated tweet filtering mechanism. By effectively removing redundant content and minimizing noise, it focuses on capturing tweets that genuinely mirror market sentiments. This optimized approach not only reduces the computational strain, which in turn allows for a more cost-effective device setup, but also enhances the accuracy of our subsequent market predictions.

\textbf{\textbullet\ Developing a self-aware mechanism for multi-level dynamic financial analysis.} 
The proposed ECON learns the dynamic and temporal distance-based relationships inherent within stocks, sectors, and macroeconomic factors through a self-aware process. 
This allows ECON to not only focus on individual stocks but also to harness effective market information by hedging against macro and micro trends.

\textbf{\textbullet\ Formulating a multi-dimensional perception of stock market in a multi-task fashion.} 
By integrating multi-dimensional correlations of stock market, we can not only predict stock price movements but also efficiently extract information from stock market volatility. By capitalizing the same day stock price movements, the model greatly amplifies its precision in predicting stock market volatility. This allows us to provide advance warning of any unusual fluctuations in the stock market in the future.

\textbf{\textbullet\  Validating the
effectiveness and efficiency of the proposed model via comprehensive experiments.} We conduct experiments on one real-world dataset\footnote{Our dataset is available at https://github.com/hao1zhao/Bigdata23.}. Both conventional methods and deep learning based methods for
stock market movements
and volatility are selected for comparisons. Evaluations of
various metrics are presented, illustrating the effectiveness of our proposed
model.

\begin{figure*}
    \centering
    \includegraphics[width=\textwidth]{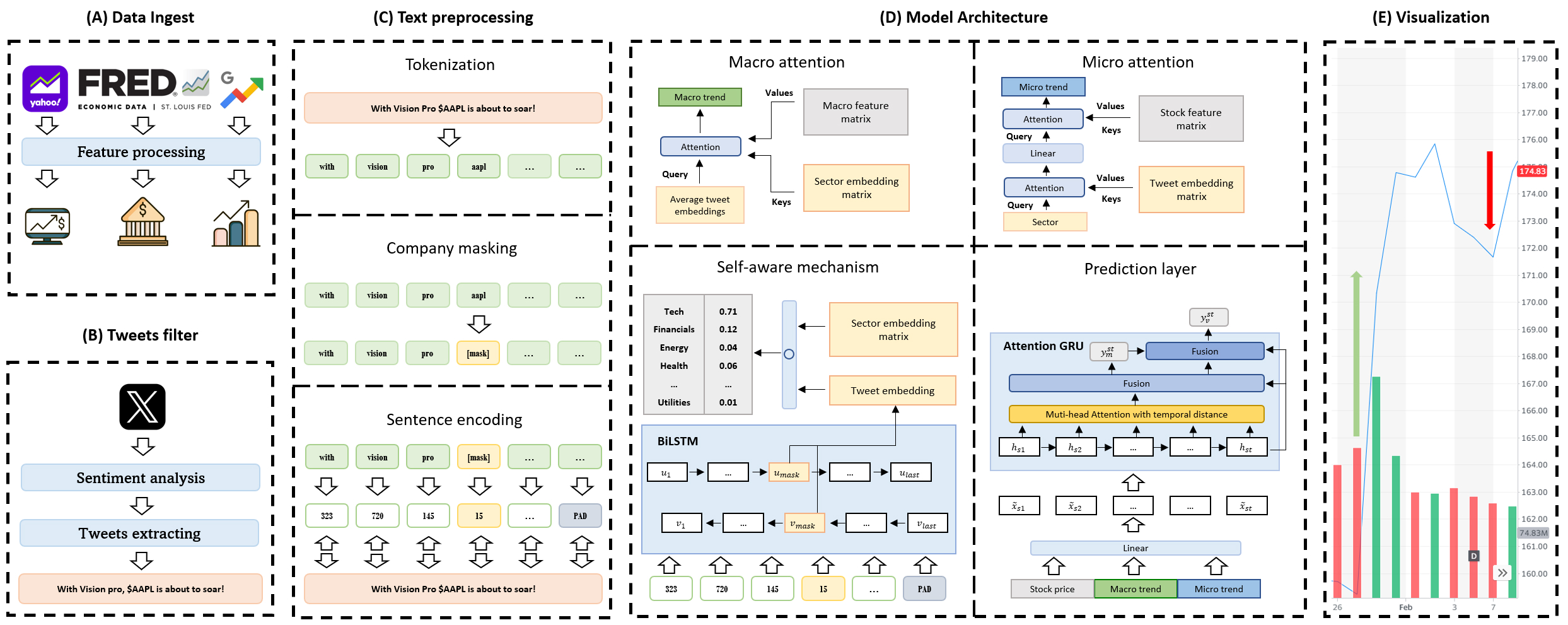}
    
    \label{fig:your_label}
    \par\vspace{5pt} 
    \parbox{\textwidth}{
        Figure 1: The architecture of ECON is designed to predict the movement \(y_m^{st}\), and volatility \(y_v^{st}\) of stock \(s\) on day \(t\). In the data input phase, information related to stock prices and macroeconomic factors is extracted. An optimal subset of tweet data is chosen via the tweet filter. Once company specifics are masked, ECON derives sector and tweet embeddings through self-aware mechanism, subsequently generating macro and micro trend features. The attention GRU amalgamates these two trends with the stock price features to make predictions.
    }
\end{figure*}
\section{Related Work}
Methods for predicting stock movement predominantly classify into two main categories: technical analysis and fundamental analysis. Technical analysis strictly hinges on historical prices to forecast future movements. In contrast, fundamental analysis casts a wider net, integrating not just historical prices but also data from textual sources, economic indicators, financial metrics, and a diverse range of both qualitative and quantitative factors.
\subsection{Technical analysis}
Technical analysis harnesses the historical prices of stocks as predictive features for future movements. The majority of contemporary methods employ deep learning architectures to unravel stock trends. Prominently, recurrent neural networks, with LSTM as a standout example, have been instrumental in identifying the temporal dynamics of stock prices \cite{choi2018stock}. The integration of temporal attention mechanisms has further enhanced the effectiveness of RNN-based models by amalgamating information from multiple time points \cite{feng2018enhancing}. Moreover, other sophisticated neural frameworks, such as convolutional neural networks and the Distillation model \cite{tang2020add}, have demonstrated their aptitude in capturing the intricate non-linear behaviors inherent to stock prices.

A significant drawback of these methods is their inability to forecast price movements when the necessary information is absent from historical prices. In this paper, our foremost objective is to surpass these technical strategies by adeptly integrating tweet and macroeconomic data into our model.
\subsection{Fundamental analysis}
It is widely believed that information sources such as tweets, or government reports give meaningful evidence for stock price prediction. Most fundamental analysis studies concerning the financial market utilize text information for stock price prediction, often employing data sources such as social media, news articles, web search queries, or government documents.

\textbf{Social Media}:
Various studies have ventured into predicting stock market leveraging social media data. Recent research employs sentiment analysis to fuse historical price data with insights gathered from platforms like Yahoo's message board \cite{das2001yahoo} , blogs\cite{de2008can}, Twitter\cite{xu-cohen-2018-stock}, and Reddit\cite{long2023just}, subsequently uncovering correlations with stock market trends.

\textbf{News and Search Engine}:
Research has delved into the intersection of public news and user browsing activity, investigating their influence on traders' reactions to news events. Studies like those by Xiong et al.\cite{xiong2015deep} view Google trends and market data as drivers behind daily S\&P 500 fluctuations. Other research, such as that by Bordino et al.\cite{bordino2012web}, correlates search query traffic with stock movements. Advanced methods, such as the hierarchical attention mechanisms proposed by Hu et al.\cite{hu2018listening}, mine news sequences directly from texts for stock trend prediction.

\textbf{Macroeconomic Indicators}:
Studies have identified a multitude of economic factors that contribute to the determination of stock returns. Particularly, Ferson et al. \cite{ferson1991variation} demonstrates that interest rates play a pivotal role in shaping stock returns. Furthermore, Jank et al. \cite{jank2012mutual} suggests additional indicators like the relative T-Bill rate and the consumption-wealth ratio to be significant predictors. Besides these factors, economic indicators such as unemployment rates, inflation, and commodity prices have also been proven to notably influence stock returns\cite{kehoe2019asset}.

However, previous studies utilizing social network data present two significant limitations. First, their predictions of stock movements concentrate solely on the company, overlooking the broader context of its industry, even though the two are deeply interconnected. Second, these studies do not incorporate a reliable information filtering mechanism, making them susceptible to the plethora of misinformation prevalent on social platforms and public media. This over-reliance can compromise the accuracy of stock movement predictions.

In this study, we aim to overcome the limitations of previous methodologies. We adopt an efficient tweet filter to eliminate redundant content and reduce noise. Additionally, we embrace a multi-dimensional approach to the stock market, integrating correlations of stock market information in a multi-task manner. By skillfully combining the strengths of both technical and model, we achieve state-of-the-art performance in predicting stock movements.

\section{Problem Setup}
We offer a formal definition for the problem of predicting stock movement and volatility.  Let \( S \) denote a set of target stocks for which predictions are sought.  We define a set \({x_{st}}\) of feature vectors, with \( s \in S \) and \( t \in T \), which encapsulate historical prices. Let \( T \) represents the set of available training days. In addition, we identify a set \( \varepsilon \) consisting of tweets, each referencing at least one stock in $S$. These historical prices incorporate the opening, highest, lowest, closing prices and adjusted closing price\footnoterule\footnote{The adjusted closing price amends a stock's closing price to reflect that stock's value after accounting for any corporate actions. } for each stock. Using $P$ to signify the adjusted closing price of stock $s$ on day $t$ as $P = \{P_{s,1}, \ldots, P_{st}\}$, we then express the actual labels for movement and volatility as follows:
\begin{equation}
\hat y_m^{st} = \mathbb{1}(P_{st} - P_{s,t-1}),
\end{equation}
\begin{equation}
\hat y_v^{st} = \mathbb{1}(P_{st} - P_{s,t-1})/P_{s,t-1}.
\end{equation}
The crux of our problem lies in forecasting the binary movement and volatility of each stock's price on day \( t  \), using the associated features and tweets within the lagged timeframe  \( [t- d , t - 1]\), where $d$ represents a predefined window size.

We usually can generate multiple training examples by shifting the time lag within extensive historical stock data in practical situations. However, in order to simplify the explanation of our proposed method, we focus on a specific time lag for both movement and volatility prediction.
We've also integrated the use of adjusted stock prices into our predictive models.\cite{xie2013semantic}. This allows our models to learn from the genuine changes in a stock's value that occur due to market conditions, rather than being swayed by artificial fluctuations resulting from corporate actions.

\section{ECON Structure}
We present an overview of data ingestion, text preprocessing, and model architecture, as illustrated in Figure 1. Initially, we process the historical stock prices and macroeconomic factors to distill pertinent features. Following this, we sift through Twitter data to identify tweets that most aptly reflect public sentiment. During text preprocessing, we mask stock tickers associated with companies found in the curated tweets. These masked tweets are then transformed into embeddings. Leveraging a self-aware mechanism, we generate embeddings for both sectors and tweets. In the concluding step, we amalgamate information from stocks, sectors, and macroeconomic indicators in a multi-trends approach to predict stock price movement and volatility.

\subsection{Data Ingest} 
\textbf{Yahoo Finance}. Yahoo Finance is a comprehensive financial news and data platform providing stock market information, financial reports, and investment resources. We obtained historical data from Yahoo Finance, tracking the performance of 42 blue-chip stocks from June 1, 2020, to June 1, 2023. To streamline our prediction targets, we set a threshold range from -0.5\% to 0.5\% to exclude minor fluctuations, as in recent works for stock movement prediction\cite{xu-cohen-2018-stock}, \cite{soun2022accurate}. According to Baker et al.\cite{baker2021triggers}, daily stock price changes above 2.5\% are often considered significant. However, our model is designed to predict unusual volatility. Therefore, in line with  Ding et al.'s \cite{ding2007private} explanation on special stock fluctuation constraints, we've set a higher benchmark, identifying a 5\% fluctuation as an abnormal change. Consequently, we label samples with fluctuation percentages less than 5\% as 0 and those equal to or more than 5\% as 1.

\textbf{Google Trends}. Google Trends is a tool offered by Google, providing users with a normalized count of total searches for specific terms within a given period. In this system, the volume of searches is benchmarked against a normalized scale, where 100 represents the peak search interest and 0 the lowest. For our study, we sourced our search data from web searches. Drawing inspiration from Wikipedia's "Outline of economics" page\cite{wikiEconomics}, we curated and refined our keyword list based on expert knowledge. Given the weekly granularity of our data, we partitioned our data retrieval into distinct time windows. We then normalized the search volumes within each window against each other, ensuring consistent normalization over the entire three-year span. This approach allowed us to seamlessly extract macroeconomic-specific trends by querying the chosen keywords.

\textbf{Federal Reserve Economic Data (FRED)}. The FRED database, overseen by the Federal Reserve Bank of St. Louis, contains over 816,000 economic time series from diverse sources, encompassing sectors like banking, employment, GDP, and more. Many of these series are gathered from government entities like the U.S. Census and the Bureau of Labor Statistics. For our study, we used the same macroeconomic keywords as in Google Trends, aiming to capture insights from both societal and official viewpoints.

\textbf{Twitter}. Twitter is a social media platform where users share and interact with brief messages called ``tweets''. During the same date range with Yahoo Finance data, we included 7.7 million tweets, gathered via Twitter's official API at a sampling rate of 10\%. Our criteria for tweet selection was stringent: they needed to contain at least one cashtag (\$) and had to be posted within standard U.S. trading hours. Besides, we recognized the significant influence that Twitter volume has on stock trading, a fact underscored by Cazzoli et al. \cite{cazzoli2016large}. Thus, we ensured that our stock feature matrix included the daily count of processed Twitter posts.

\subsection{Tweets filter} 
In stock analysis leveraging Twitter data, two predominant text processing approaches emerge due to token length constraints: the first approach uses pre-trained models to gauge tweet sentiments, converting these into sentiment scores. Consider a tweet such as ``With Vision pro, \$AAPL is about to soar!'' This would likely be identified as positive in sentiment. These scores subsequently serve as features in model analysis. However, this method's primary shortcoming lies in its exclusive emphasis on sentiment, overlooking the depth of other textual nuances. Additionally, the vast sea of collected tweets often contains a great deal of redundancy, with only a fraction truly being insightful. This leads to an analysis that may not genuinely capture the overarching public sentiment.

The second strategy employs a custom encoder to derive features from tweets, setting a daily limit on the number of tweets analyzed for each stock, often not surpassing a few hundred tokens. At a glance, this may seem practical. However, given the vast number of tweets we encounter daily, often averaging in the hundreds with surges nearing 10,000, this approach becomes increasingly untenable. Relying on a limited, randomly chosen subset of tweets to extract influential features not only tests our analytical prowess but also brings into question the breadth and validity of the conclusions drawn.

To tackle the challenges previously outlined, we developed a sophisticated tweet filtering system. Each tweet's sentiment is ascertained via sentiment analysis. For stock $s$ on day $t$, tweets are prioritized according to their impressions. The number of tweets retained daily is determined by assessing the correlation between Sentiment and Movement, leveraging Cramer’s V coefficient $V$. In our research, we consistently selected the top six tweets, those garnering the highest daily impressions, as our primary input for text preprocessing.

\textbf{Sentiment Analysis}.
In our research, we undertook a sentiment analysis on a dataset comprising 7.7 million tweets. To achieve this, we leveraged a language model based on Roberta\footnoterule\footnote{https://huggingface.co/cardiffnlp/twitter-roberta-base-sentiment-latest.}, and the model has been fine-tuned on Twitter data. This optimized model can produce one of three sentiment outcomes for each tweet: positive, neutral, or negative.
For every tweet \( e \) associated with a blue-chip stock \( s \), we compared the predicted sentiment \( p_e \) with the same day's binary adjusted price movement \( y^s_m \) of stock \( s \). The measure \( y^s_m \) gauges the price fluctuation between days \( t -1 \) and \( t  \). Same as problem setup, if the stock price ascends in this interval, \( y^s_m \) is labeled ``Up'', and conversely, if it declines, it's labeled ``Down''. It's pertinent to note that our dataset only incorporated tweets from the trading hours of day \( t \).
To aggregate daily sentiment, if a day witnesses a predominance of negative sentiments in tweets concerning a specific stock, we designate that day as ``negative''. In contrast, if the tweets are primarily characterized by positive sentiments, the day is classified as ``positive''.

\begin{table}[ht]
\parbox{\columnwidth}{
\normalsize
    TABLE I : Observed frequency between stock sentiment and stock price movement.
    \par\vspace{5pt} 
}
\centering
\resizebox{\columnwidth}{!}{%
\begin{tabular}{|c|c|c|c|c|}
\hline
          & $Movement_1$ & $Movement_2$ & ... & $Movement_j$ \\ \hline
$Sentiment_1$ & $O_{11}$        & $O_{12}$        & ... & $O_{1j}$        \\ \hline
$Sentiment_2$ & $O_{21}$        & $O_{22}$        & ... & $O_{2j}$        \\ \hline
...       & ...      & ...      & ... & ...      \\ \hline
$Sentiment_i$ & $O_{i1}$        & $O_{i2}$        & ... & $O_{ij}$        \\ \hline
\end{tabular}%
}

\label{tab1}
\end{table}

\begin{figure*}[!htb]
  \centering
  \subfigure[Top one impression tweet ]{
    \includegraphics[width=0.30\textwidth]{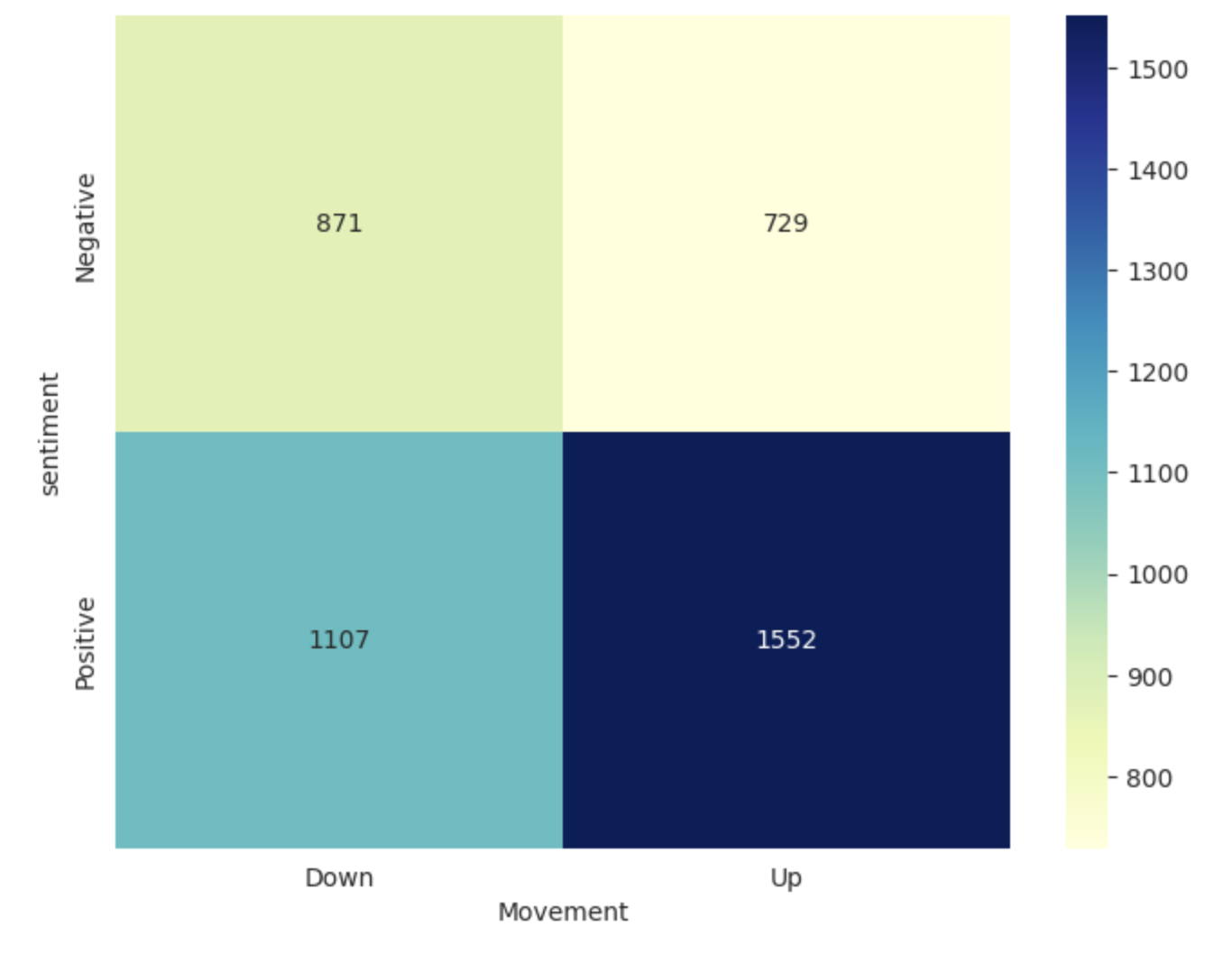}
    \label{fig:pic1}
  }
  \hspace{0.015\textwidth}
  \subfigure[Top three impression tweets ]{
    \includegraphics[width=0.30\textwidth]{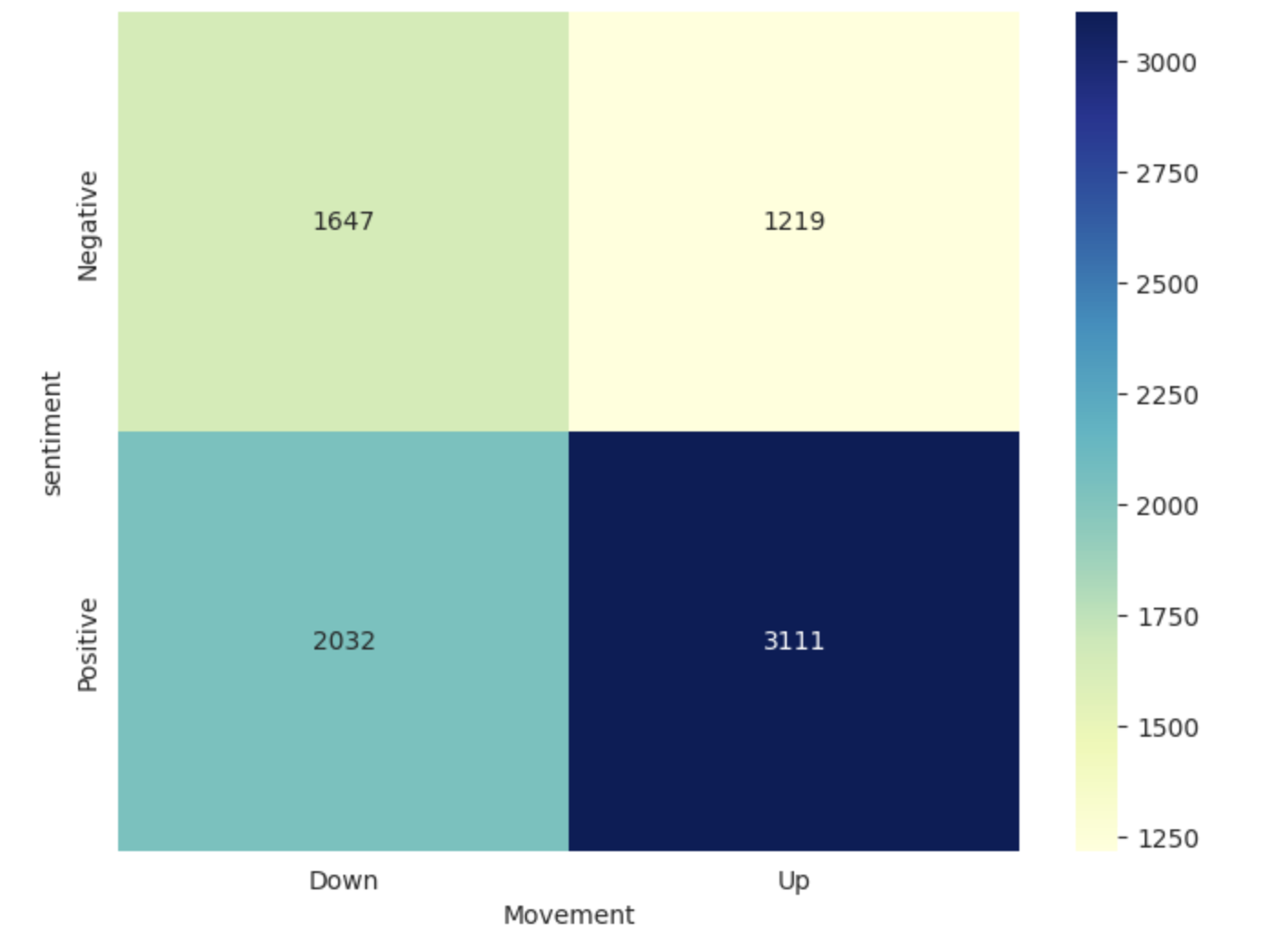}
    \label{fig:pic2}
  }
  \hspace{0.015\textwidth}
  \subfigure[Top six impressions tweets ]{
    \includegraphics[width=0.30\textwidth]{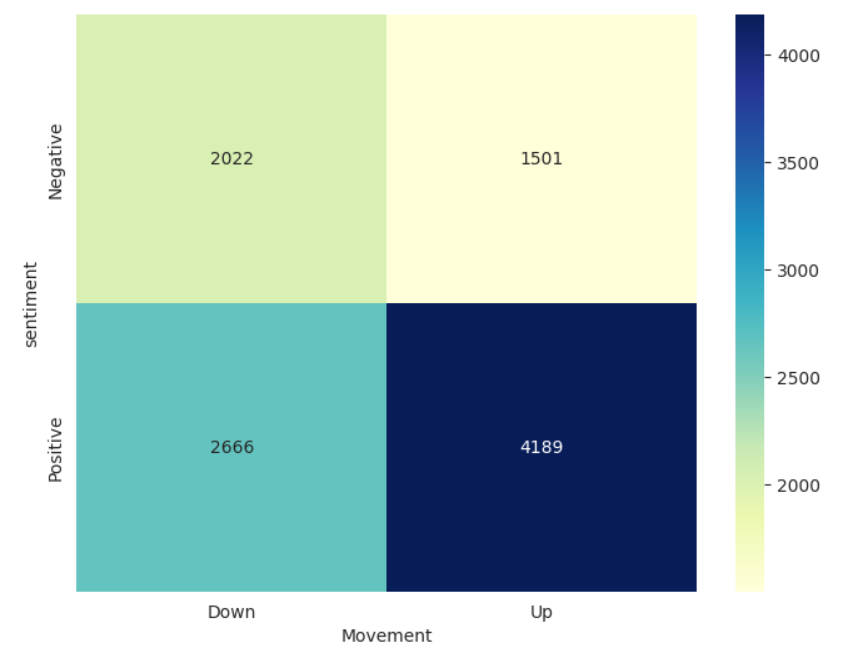}
    \label{fig:pic3}
  }
  
  \parbox{\textwidth}{
        Figure 2 : Confusion matrices for both sentiment scores and stock price movements.  (a), (b), and (c) each depict the results of selecting different numbers of tweets each day to typify that day's Twitter dataset. We can observe a clear link between stock price trends and tweet sentiments.
    }
  \label{fig:combined}
\end{figure*}

\textbf{Tweets extracting}.
In addition to textual information, we gathered metrics on likes, retweets, and impressions for each tweet. Impressions on Twitter signify the cumulative count of times a tweet appears in user timelines, search results, or other viewing sections of the platform. While Twitter remains reticent about the exact workings of its algorithm, Busch et. al \cite{6228205} explained that user actions affect how often a tweet is pushed out, referred to as its ``impressions''. This implies that the greater the impression of a tweet, the more people likely agree with its viewpoint. Therefore, using tweets with a larger impression provides a better representation of society's perspective on that stock. Due to the constraints of the actual production environment, it is hard to process thousands of tweets simultaneously. Moreover, many of these tweets are redundant, such as those without meaningful content or those that are purely promotional. Therefore, we sorted the each tweet $e$ based on their impressions $k$ and calculated the Cramér’s V coefficient $V$ between the overall sentiment of distinct daily tweets and price movement.
Through the contingency Table I, where each cell value represents an observed frequency $O_{ij}$, we can derive the expected frequencies as follows: 
\begin{equation}
E_{ij} = \frac{Row_i \times Column_j}{Total_{ij}},
\end{equation}
where \(Row_i\) represents the total count for the \(i^{th}\) sentiment category, while \(Column_i\) signifies the total count for the \(j^{th}\) movement category from the training set. \(Total_{ij}\) encompasses the sum of all values within the contingency table.

We iterated through all categories of the two variables, utilizing the following chi-square statistic function to measure the discrepancy between observed and expected frequencies:

\begin{equation}
\chi^2 = \sum_{i} \sum_{j} \frac{(O_{ij} - E_{ij})^2}{E_{ij}}.
\end{equation}

As shown in Figure 2, there is a discernible correlation between Movement and Sentiment. In our study, for stock $s$ on day $t$, we sorted each tweet $e$ based on its impression count. Subsequently, we sequentially extracted distinct tweets to construct multiple datasets $m$. By comparing the Cramér's V for Sentiment and Movement across these datasets, we ascertained their correlation, as presented in Table II. The formula for the Cramér's V is provided as follows:

\begin{equation}
V = \sqrt{\frac{\chi^2}{Total_{ij} \times (min(Row_i,Column_j) - 1)}}.
\end{equation}

Upon comparison, we observed that the Cramér's V for datasets $m_k$ containing all tweets from the training set is slightly higher. However, incorporating all daily tweets into the model in practical applications presents a significant computational overhead. Therefore, we opted to select the top six tweets $m_6$ for each stock daily as our finial data input, given their commendable Cramér's V on a smaller scale.

\begin{table}[ht]
\centering
\parbox{\columnwidth}{
\normalsize
    TABLE II : A varying number of daily tweets are retained for comparison using Cramér's V. For $m_1$, only the tweet with the highest impression each day is retained. Following this pattern, $m_k$  represents the retention of all tweets. 
    \par\vspace{5pt} 
}
\Large                  
\renewcommand{\arraystretch}{1.5} 
\resizebox{\columnwidth}{!}{%
\begin{tabular}{|c|c|c|c|c|c|c|c|c|c|c|}
\hline
  & $m_1$      & $m_2$      & $m_3$      & $m_4$      & $m_5$      & $m_6$      & $m_7$      & $m_8$      & ... & $m_k$      \\ \hline
$V$ & 0.0394 & 0.0643 & 0.0747 & 0.0816 & 0.0833 & 0.0877 & 0.0872 & 0.0853 & ... & 0.0917 \\ \hline
\end{tabular}%
}
\label{tab2}
\end{table}

\subsection{Text preprocessing}
In natural language processing (NLP), text preprocessing has long been recognized as a key step. By streamlining text into a format more amenable to training, it not only simplifies the content but also boosts the efficiency of machine learning algorithms. Shifting focus to the concept of self-aware mechanism, our main objective is pinpointing sector identification. Within this context, the embeddings of both sectors and tweets are crucial. When the stock symbol $s$ is masked, we aim to identify the sector $c$ a tweet $e$ refers to based solely on its content.

\textbf{Tokenization}. Tokenization is the process by which text is divided into smaller units, such as sentences, words, characters, or subwords. Word tokenization, specifically, refers to the segmentation of text into individual words. In our research, we utilized spaCy\footnoterule\footnote{https://spacy.io/}—an open-source natural language processing tool—for word tokenization of every tweet $e$. Moreover, a stock ticker like ``AAPL'' for Apple might inadvertently be tokenized into ``AA'' and ``PL''. To circumvent such issues, we established specific rules to ensure that stock tickers are consistently recognized as single, intact tokens.  After tokenization, words are converted to lowercase because stock tickers won't overlap with regular vocabulary. Such normalization not only ensures vocabulary consistency across the dataset but also augments computational efficiency.

\textbf{Company Masking}.
For the subsequent self-awareness portion of the model architecture, we replace all tokens in $\varepsilon_s$ that correspond to the name of stock s with a special token [mask]. Based on the sector classification rules of GISC, we maintained a dictionary mapping the masked companies to their respective sectors. For example, after data filtering, we obtain a tweet with a high impression: ``With Vision pro, \$AAPL is about to soar!'' Next, we mask the stock ticker, converting it to: ``With Vision pro, [mask] is about to soar!''

\textbf{Sentence encoding}. Words from the sentence are first tokenized and then mapped to their corresponding word embeddings. These embeddings are high-dimensional vectors that capture the intrinsic semantics of each word. We record the original length of each tweet, storing this information in the variable $l$. This data proves crucial when padding comes into play, a technique used to append or prepend sequences with filler values, ensuring uniformity. Given the importance of maintaining consistent sequence lengths, especially when working in batches, padding ensures our dataset's sequences adhere to a standardized length, adding consistency and structure to our data processing pipeline.

\subsection{Model Architecture}
For blue-chip companies, the categorization into sectors is primarily based on their core business operations. However, given the substantial scale of these enterprises, their business models are inherently diverse and not confined to a single domain. In fact, many such corporations have a presence across multiple industries. Take Apple as an example: its robust cash flow and stable user base have increasingly amplified its influence in the financial sector. The rapid growth of Apple Card's high-yield savings account\footnoterule\footnote{The saving account jointly launched by Apple and Goldman Sachs, offering an annual percentage yield significantly above the national average. }, amassing over \$10 billion in deposits from users within just four months since its launch in April 2023, serves as a testament to this diversification. On the other hand, The rise and fall of a company's stock price are often influenced by its respective sector. Whenever a significant event occurs, the impact varies across different sectors\cite{he2020covid}.

\textbf{Self-aware mechanism}.
Our methodology for modeling sector and tweet embeddings takes two distinct approaches. For sectors, the embedding $h_c$ of each sector $c$ is learned as a free parameter, updated dynamically during backpropagation. Each embedding $h_c$ is the c-th row of the sector embedding matrix $C$. Conversely, for tweets, in line with our earlier discussions on company masking, we then deploy a predictor network and harness its output $f(e)$ based on the tokenized content of each tweet $e$ to derive the embedding $h_e$.

Given the intrinsic capability of recurrent neural networks to effectively manage long-term dependencies, they are a preferred choice for sequential data processing, as evidenced by various literature~\cite{xu2018stock}. In our approach, we adopted a bi-directional LSTM (BiLSTM) for the predictor network  to produce the tweet and sector embeddings. The reason why we use BiLSTM is because LSTM cannot simultaneously account for the context on both sides of the masked token—a key requirement for our sector identification task. To be more precise, we execute the BiLSTM and utilize the state vector produced at the masked token position as $h_e$. This ensures our model places emphasis on the sector context surrounding the masked token rather than the end context of the tweet.

We train three distinct sets of parameters: the word token embeddings, the parameters within the predictor network, and the sector embeddings. Utilizing the tweet embedding, denoted as $h_e$, generated by the predictor network, we define our prediction for the sector $c$ as:

\begin{equation}
{y_{ec}} = \frac{\exp(h_c^T h_e)}{\sum_{c' \in C} \exp(h_{c'}^T h_e)},
\end{equation}
where $c'$ represents all possible sectors within the set $C$. For our dataset, we can categorize it into 10 distinct sectors.

The sector self aware training is done separately from the predictions section. To optimize our model, we adjust all parameters to minimize an objective function derived from the cross entropy loss, detailed as follows:

\begin{equation}
l(\theta) = - \sum_{c \in C} \sum_{e \in \varepsilon_s} \log{{y_{ec}}}.
\end{equation}

As highlighted by Soun et. al\cite{soun2022accurate}, maximizing the dot product between sector and tweet embeddings during the self-aware phase allows for these embeddings to be seamlessly interchanged within the primary model designed for stock movement prediction. Our expectations from the sector self-aware mechanism are twofold: Firstly, the tweet embedding, while principally revealing details about its respective company, also imbibes information from the sector embedding under which multiple companies fall. This allows for a richer contextual understanding of the company in the wider industry setting. Secondly, Blue-chip stocks, given their expansive reach across various sectors, possess unique sector embeddings. By virtue of these companies operating in diverse sectors, amalgamating information from these sectors not only offers a more comprehensive view but can also enhance the accuracy of our predictions.

\textbf{Macro attention}.
Given that stock price movements are influenced not only by their respective sectors but also by overarching macroeconomic market factors, it becomes imperative to craft a comprehensive market summary for more accurate predictions.  

In our model, a comprehensive market index is constructed by amalgamating price-related information from diverse sectors and incorporating pertinent macroeconomic factors. This synthesis results in the macro trend vector \(a_t\), which encapsulates holistic market data for subsequent predictions. Given embedding vectors \(h_c\) for each sector \(c\) and \(h_e\) for every tweet \(e\), we derive the daily average tweet embeddings, denoted by \(r_t \in \mathbb{R}^{2k}\). Utilizing \(r_t\) as the query vector, the attention mechanism is employed to integrate price-related attributes across sectors using their specific embeddings. The specifics of the attention process are detailed as follows:

\begin{equation}
a_t = \sum_{c \in C} \alpha_{ct} x_{ct} \quad \text{where} \quad \alpha_{ct} = \frac{\exp(r^T_t h_c)}{\sum_{c' \in C} \exp(r^T_t h_{c'})},
\end{equation}
where \( x_{ct} \in \mathbb{R}^{m \times q} \) represents the macroeconomic and price-related feature of sector \( c \) on day \( t \), and \( \alpha_{ct} \) serves as the attention weight for aggregation.
Figure 1 macro attention part depicts the Macro trend derivation process. For each day \( t \), the mean tweet embedding \( r_t  \), is computed from all tweet embeddings. This average embedding serves as the query in the attention mechanism, while the matrix \( C \in \mathbb{R}^{m \times 2k}\) representing sector embeddings acts as the key. The attention mechanism's outcome is an aggregated macro feature matrix, which is comprised of data from FRED and Google Trends. Then it yields the macro trend vector \( a_t \in \mathbb{R}^{q}\). In the subsequent predictive phase, this generated vector is channeled into the Attention GRU. This provides both macro-level perspectives and a comprehensive understanding of the sector in which the company is positioned.

\textbf{Micro attention}.
Many blue-chip companies operate across multiple sectors. This diversification allows them to hedge against downturns in one sector by benefiting from the rise in another. Hence, stock price movements and volatility for these companies cannot be solely determined by macro trends. We propose to make a micro trend vector $i_{ct}$ for each sector $c$ at day $t$, which combines the stock information and relevant sectors. The micro trend \(i_{ct}\) provides cross-sectoral involvement information for each stock \(s\), which is distinct from the Macro trend vector \(a_t\) that takes into account the influences of macroeconomics and societal hot topics for all target sectors. The first attention is to find tweets relevant to each target sector $c$. We find relevant tweets by using the embedding $h_c$ of sector $c$ as the query of attention as follows:
\begin{equation}
w_{ct} = \sum_{e \in \varepsilon_t} \alpha_e h_e \quad \text{where} \quad \alpha_e = \frac{\exp(h^T_c h_e)}{\sum_{e' \in \varepsilon_t} \exp(h^T_c h_{e'})}.
\end{equation}

Considering a day \( t \), let \( \varepsilon_t \) represent the collection of tweets. For each tweet \( e \) within this set, its corresponding embedding is denoted by \( h_e \). Additionally, \( \alpha_e \) signifies the attention weight attributed to the tweet. The vector \( w_{ct}\in \mathbb{R}^{2k}\), which can be envisioned as a weighted mean of all the tweet embeddings for day \( t \). This averaging is influenced by the 
pertinence of each tweet to the specific target sector \( c \).
Then, we employ a linear layer to project $w_{ct} $ onto all possible companies $S$:
\begin{equation}
g_{st} = W_1(w_{ct}) + b_1,
\end{equation}
where $W_1 \in \mathbb{R}^{n \times 2k}$, $b_1 \in \mathbb{R}^{n}$. As a result of the attention function of Equation (10),  $g_{st} \in \mathbb{R}^{n}$ represent that summarizes all tweets at day $t$ considering the relevance to sector $c$. Our second attention uses the resulting tweet vector to find the most relevant stocks at the moment. This is done by using $g_{st}$ as the query vector of attention to aggregate the stock price features at day $t$ based on the stock embeddings:
\begin{equation}
i_{st} = \sum_{s \in S} \alpha_{st} x_{t} \quad \text{where} \quad \alpha_{st} = \frac{\exp(g^T_{st} x_t)}{\sum_{s' \in S} \exp(g^T_{st} x_{t'})}.
\end{equation}

The micro trend vector \(i_{st} \in \mathbb{R}^{n}\) encapsulates all tweets for day \(t\), integrating the information from companies associated with the sector. This vector is then combined with the stock price vector \(x_{t} \in \mathbb{R}^{n \times p}\) which generates from Yahoo's stock data and the macro trend vector \(a_t\), serving as the input to the Attention GRU for forecasting stock movement and volatility.

\textbf{Predictions layer}. 
The model is primarily designed to extract nuanced information from stock market, macroeconomic factors, and tweet data. Once the macro trend \(a_{t}\) and micro trend \(i_{st}\) are generated, they are concatenated with the stock price feature \(x_{st} \in \mathbb{R}^{p}\) using a linear layer, producing a multi-level feature. The corresponding function is as follows:

\begin{equation}
\tilde{x}_{st} = W_2(x_{st} \oplus a_{t} \oplus i_{st}) + b_2,
\end{equation}
where $W_2 \in \mathbb{R}^{n \times (p+q+n)}$, $b_1 \in \mathbb{R}^{n}$ are the learnable weight and bias terms, respectively, $\oplus$ is the concatenation operator between vectors.

\textit {Attention GRU with temporal distance (AGRUD).} Given the proficiency in handling long-term dependencies, recurrent neural network is extensively employed for sequential data processing~\cite{xu2018stock}. The general idea of recurrent unit is to recurrently project the input sequence into a sequence of hidden representations. At each time-step for stock $s$, the recurrent unit learns the hidden representations \(h_{st}\) by jointly considering the input \(\tilde{x}_{st}\) and previous hidden representation \(h_{s,t-1}\) to capture sequential dependency. To capture the sequential dependencies and temporal patterns in the historical stock features, an GRU recurrent unit\cite{cho2014learning} is applied to map $ \{\tilde{x}_{s1}, \ldots, \tilde{x}_{st}\}$  into hidden representations $ \{h_{s1}, \ldots, h_{st}\}$. Then, We implement a distance matrix to apply weights to the hidden states of the GRU:
\begin{equation}
h_{st}' = h_{st} \times \frac{1}{\Delta d},
\end{equation}
where $\Delta d$ represents the distance between a specific day within the window $d$ and the day $t$. This approach ensures that hidden states further away from the current time step receive progressively lower weights, emphasizing the relevance of more recent states. Then, the states are combined by attention as follows:


\begin{equation}
{h}_{att} = \sum_{s=1}^{S} \sum_{t=1}^{T} \alpha_{st} h'_{st} \quad \text{where} \quad \alpha_{st} = \frac{\mathbf{u}^T h'_{st}}{\sum_{s'=1}^{S} \sum_{t'=1}^{T} \mathbf{u}^T h'_{s't'}},
\end{equation}
in this context, $u$ is a learnable parameter, commonly referred to as the attention query. It serves to select the most pertinent time steps by evaluating the outcome of the dot product. The derived weight $\alpha_{st}$ indicates the degree to which step $t$  contributes to $h_{att}$.

Attention GRU with temporal distance generates the first output $h_{out_1} = h_{st}' \oplus h_{att}$ by concatenating the hidden state $ h_{st}'$ of the last time step and the output $h_{att}$ of the attention, where $\oplus$ is the concatenation operator between vectors. The last hidden state $h_{st}'$ is used in addition to $h_{att}$ as the basic output that summarizes all given features apart from the result of attention. Building upon the previous output $h_{out_1}$, we further concatenate $h_{att}$ and $h_{st}'$ to produce the second output $h_{out_2} = h_{out_1} \oplus h_{att} \oplus h_{st}'$.

We utilize the attention GRU with temporal distance, as introduced above, as our primary tool for forecasting stock movement and volatility. The model take the output $\tilde{x}_{st}$ from the linear layer as its input. The subsequent formula outlines the methodology for predicting stock price movements:

\begin{equation}
y_v^{st}, y_m^{st} = \text{AGRUD}(\tilde{x}_{s,t-d}, \tilde{x}_{s,t-d+1}, \dots, \tilde{x}_{st}),
\end{equation}
where $d$ is the window size, which is chosen as a hyperparameter between 5 and 15 in our experiments.

\textit {Optimization.} We update all parameters to minimize the following objective function, which is based on the cross entropy function for stock price movement:

\begin{equation}
l(\theta) = - \sum_{s \in S} \sum_{t \in T} \log{{y_m^{st}}},
\end{equation}
where $\theta$ is the set of learnable parameters of AGRUD, $T$ is the set of available days in training data , $S$ is the set of target stocks.

We also train AGRUD to minimize the following objective function for stock price volatility: 
\begin{equation}
L(\theta) = - \sum_{s \in S} \sum_{t \in T}\left[\hat y_v^{st} \log(\sigma( y_v^{st})) + (1 - \hat y_v^{st} ) \log(1 - \sigma( y_v^{st})) \right]
\end{equation}
where $\hat y_v^{st} \in \{0,1\}$  is the true label of stock $s$ at day $t$. As previously mentioned, the true label for stock price volatility is expounded upon in equation (2).

\section{Experiment}

We conduct experiments to answer the following questions about the performance of framework:

\noindent Q1.  \textbf{Baseline methods}. How do other baseline models perform on our task? 

\noindent Q2. \textbf{Tweets quality}. Does using our model to extract data yield better results than performing sentiment analysis using the entire 7.7 million tweets as input?

\noindent Q3. \textbf{Ablation study}. How do the macro and micro trend affect its performance?

\begin{table}[ht]
\centering
\parbox{\columnwidth}{
\normalsize
    TABLE III : Classification performance of ECON and baseline methods, measured with the accuracy (ACC) , the Matthews correlation coefficient (MCC) and Receiver Operating Characteristic (ROC). The best is in bold, Our ECON shows the best performance in all  evaluation metrics.
    \par\vspace{5pt} 
}
\label{tab3}
\begin{tabularx}{\columnwidth}{@{}ll*{5}{>{\centering\arraybackslash}X}@{}}
\toprule
\multicolumn{2}{c}{\multirow{2}{*}{Models}} &
  \multicolumn{2}{c}{Movement} &
  \multicolumn{3}{c}{Volatility} \\ 
\cmidrule(r){3-4} \cmidrule(r){5-7} 
\multicolumn{2}{c}{} &
  Acc. & 
  MCC &
  Acc. &
  MCC &
  AUC \\ 
\midrule
XGBoost & & 49.30 & 0.0146 & 52.40 & -0.0127 & 55.61 \\
ARIMA & & 50.71 & 0.0031 & 53.61 & -0.0032 & 56.89 \\
ALSTM  & & 50.22 & 0.0453 & 56.43 & 0.0624 & 59.58 \\
Adv-LSTM & & 51.76 & 0.0374 & 58.84 & 0.0529 & 58.93 \\
DTML & & 51.95 & 0.0578 & 59.79 & 0.0703 & 61.06 \\
SLOT & & 52.45 & 0.0617 & 60.83 & 0.0628 & 60.52 \\
ECON & & \textbf{53.36} & \textbf{0.0754} & \textbf{62.84} & \textbf{0.0861} & \textbf{63.54}\\ 
\bottomrule
\end{tabularx}
\end{table}

\subsection{Baseline Methods}
We evaluate our model's effectiveness by comparing it with technical and fundamental analysis models. Technical analysis models focus on capturing patterns of historical prices and fundamental analysis models combine historical prices with other government and financial market information for stock price movement and volatility prediction.

\noindent\textbullet \ \textbf{Extreme Gradient Boosting (XGBoost)}~\cite{chen2016xgboost} is a high-performance gradient boosting framework renowned for its efficiency in handling large datasets, regularization features, and capability in both classification and regression tasks.

\noindent\textbullet \ \textbf{Autoregressive Integrated Moving (ARIMA)}~\cite{brown2004smoothing}
is a forecasting model for time series data, consisting of autoregressive (AR), differencing (I), and moving average (MA) components. It assumes data linearity and stationarity.

\noindent\textbullet \ \textbf{Long Short-Term Memory (LSTM)}~\cite{hochreiter1997long} is a type of recurrent neural network (RNN) architecture specifically designed to tackle the vanishing and exploding gradient problems encountered in traditional RNNs. LSTM units use gate mechanisms to regulate the flow of information, making them well-suited for learning from long-term dependencies in sequences.

\noindent\textbullet \ \textbf{Attention LSTM (ALSTM)}\cite{feng2018enhancing} is a LSTM network with an attention mechanism, enhancing sequence modeling by emphasizing relevant elements.

\noindent\textbullet \ \textbf{DTML}\cite{yoo2021accurate} is based on Transformer, it learns temporal correlations and combines the contexts of all target stocks.

\noindent\textbullet \ \textbf{SLOT}\cite{soun2022accurate} is based on ALSTM, it captures correlations between tweets and stocks through self-supervised learning.

To test whether our Twitter filter can still accurately capture market sentiment even with a significantly reduced Twitter input:

\noindent\textbullet \ \textbf{AGRUD-A}  concatenated the sentiment score of all 7.7 million tweets with stock prices and macroeconomic data, keeping the structure of main predictor AGRUD unchanged. 

\noindent\textbullet \ \textbf{AGRUD-F}  concatenated the sentiment score of data refined  with stock prices and macroeconomic data, keeping the structure of main predictor AGRUD unchanged.

\begin{table}[ht]
\centering
\parbox{\columnwidth}{
\normalsize
    TABLE IV : This table illustrates the effectiveness of the tweet filter. By retaining only the tweets with the highest impressions, the results for predicting stock price movement and volatility are consistent with those obtained using sentiment analysis on all tweets.
    \par\vspace{5pt} 
}

\label{4}
\begin{tabularx}{\columnwidth}{@{}l*{6}{>{\centering\arraybackslash}X}@{}}
\toprule
\multicolumn{1}{c}{\multirow{2}{*}{Models}} & \multirow{2}{*}{Tweets} &
  \multicolumn{2}{c}{Movement} &
  \multicolumn{3}{c}{Volatility} \\ 
\cmidrule(r){3-4} \cmidrule(r){5-7} 
& &
  Acc. & 
  MCC &
  Acc. &
  MCC &
  AUC \\ 
\midrule
AGRUD-A & 7,765,542 & 52.38 & 0.0518 & 61.36 & 0.0747 & 61.97 \\
AGRUD-F & 193,147 & 52.21 & 0.0564 & 60.68 & 0.0694& 59.93 \\
ECON & 193,147 & \textbf{53.36} & \textbf{0.0754} & \textbf{62.84} & \textbf{0.0861} & \textbf{63.54} \\
\bottomrule
\end{tabularx}
\end{table}

\textbf{Evaluation metrics}. Following previous work for stock movement prediction, we adopt the standard measure of Accuracy (ACC) and Matthews Correlation Coefficient (MCC) as evaluation metrics. MCC avoids bias due to data skew. Given the confusion matrix containing the number of samples classified as true positive $(tp)$, false positive $(fp)$, true negative $(tn)$ and false negative $(fn)$, Acc. and MCC are calculated as follows:
\begin{equation}
Acc. = \frac{tp + tn}{tp + tn + fp + fn},
\end{equation}
\begin{equation}
MCC = \frac{tp \times tn - fp \times fn}{\sqrt{(tp + fp)(tp + fn)(tn + fp)(tn + fn)}}.
\end{equation}

For stock volatility prediction, given that data points involving stock price changes greater than 5\% only constitute a minor portion of our dataset, using accuracy to evaluate ECON can be misleading due to model overfitting. When evaluating the imbalanced data, we use Area Under the Curve - Receiver Operating Characteristic (AUC-ROC) where AUC is the area under the ROC curve, which falls in the range [0,1]. AUC is insensitive to the absolute numbers of positive and negative samples. It primarily focuses on how the model distinguishes between positive and negative samples. AUC is the area under the ROC curve, which falls in the range [0,1]. AUC is calculated as follows:
\begin{equation}
{AUC} = \int_{0}^{1} TPR(FPR) \, dFPR,
\end{equation}
where $TPR(FPR)$ represents the True Positive Rate  as a function of the False Positive Rate. The $TPR$ gives the proportion of actual positives that are correctly identified. $FPR$ is the False Positive Rate, indicating the proportion of negative samples that are mistakenly identified as positive.

\subsection{Results}
The performances of our ECON and the established baselines are detailed in Table III. ECON is observed to be the superior baseline model in terms of accuracy and MCC  for movement prediction. While DTML shows an outstanding performance in the AUC and MCC for volatility prediction. ECON surpasses both of these models by significant margins. In terms of accuracy, ECON achieves a score of 53.36 and 62.84, outperforming SLOT and DTML by 1.7\% and 2.7\%  respectively, and outshines in AUC by a margin of 4.0\% and 4.0\%. We argue this improvement of price movement prediction is attributed to use of macro and micro trends to investigate the influence of multiple sectors on stocks. On the other hand, leveraging macro insights and placing a stronger emphasis on recent information have also significantly contributed to improved volatility forecasting.

The performance results for the tweets filter can be seen in Table IV. When relying solely on sentiment analysis, even though the number of tweets significantly decreased, AGRUD-A and AGRUD-F results remained almost the same. This demonstrates that the filtered tweets effectively represent people's genuine intentions. Consequently, this allows ECON 
to achieve more accurate forecasting results at a reduced cost and with greater efficiency.

\begin{table}[ht]
\centering
\parbox{\columnwidth}{
    \normalsize 
    TABLE V : An ablation study of ECON, where ECON(W/O AI) is baseline without any tweet and macroeconomic data. ECON(A) is without the micro trend and ECON(I) is without macro trend. Both macro and micro trend improve the performance of the framework.
    \par\vspace{5pt}
}
\label{5}
\begin{tabularx}{\columnwidth}{@{}l*{5}{>{\centering\arraybackslash}X}@{}}
\toprule
\multicolumn{1}{c}{\multirow{2}{*}{Models}} &
  \multicolumn{2}{c}{Movement} &
  \multicolumn{3}{c}{Volatility} \\ 
\cmidrule(r){2-3} \cmidrule(r){4-6} 
& Acc. & MCC & Acc. & MCC & AUC \\ 
\midrule
ECON(A)      & 52.16 & 0.0513 & 57.93 & 0.0517 & 60.09 \\
ECON(I)      & 51.52 & 0.0392 & 55.58 & 0.0321 & 57.53 \\
ECON(W/O AI) & 49.71 & 0.0012 & 53.47 & 0.0399 & 55.40 \\
\midrule
ECON         & \textbf{53.36} & \textbf{0.0754} & \textbf{62.84} & \textbf{0.0861} & \textbf{63.54} \\ 
\bottomrule
\end{tabularx}
\end{table}

\subsection{Ablation Study}
An ablation study of ECON, where we have constructed three variations alongside the fully-loaded model. Each variant is specifically tailored to handle certain types of input data: ECON(A) solely relies on macro trend, ECON(I) exclusively processes micro trend while ECON(W/O AI)  omits both macro and micro trends information. As shown in Table V, our ablation study revealed that incorporating multi-dimensional perception of stock market significantly enhances the predictive capabilities of the model for stock movement and volatility. A noteworthy point is that the result of ECON(A) demonstrates a significant role of the macro trend in enhancing the accuracy of volatility prediction.

\section{Conclusion}
We present ECON, a robust framework tailored for forecasting stock movement and volatility. ECON harnesses Twitter data effectively, capitalizing on its ability to reflect market sentiment, even amidst its redundancy. This framework stands out in its capability to interpret intricate multi-level relationships pertinent to stock movement and volatility. Through a novel self-aware mechanism, ECON delves deep into both macro and micro trends. Furthermore, we evaluated ECON using a comprehensive, newly-curated dataset, and the results affirm its superior performance compared to well-established baselines.

\bibliographystyle{IEEEtran}
\bibliography{ref1}

\end{document}